\documentclass[10pt,twocolumn,letterpaper]{article}

\usepackage{cvpr}              

\usepackage{algorithm}
\usepackage{algorithmic}
\usepackage[accsupp]{axessibility}
  
\newcommand{\eq}{Eq. \eqref}
\usepackage[dvipsnames]{xcolor}

\definecolor{cvprblue}{rgb}{0.21,0.49,0.74}
\usepackage[pagebackref,breaklinks,colorlinks,citecolor=cvprblue]{hyperref}

\def\paperID{1943} 
\def\confName{CVPR}
\def\confYear{2024}

\title{Estimating Noisy Class Posterior with Part-level Labels for Noisy Label Learning}

\author{Rui Zhao$^{12}$ \and Bin Shi$^{12}$\thanks{Corresponding author} \and Jianfei Ruan$^{12}$ \and Tianze Pan$^{3}$ \and Bo Dong$^{24}$
\and
$^{1}$School of Computer Science and Technology, Xi’an Jiaotong University, China\and
$^{2}$Shaanxi Province Key Lab of Big Data Knowledge Engineering, Xi’an Jiaotong University, China\and
$^{3}$School of Physics, Xi’an Jiaotong University, China\and
$^{4}$School of Distance Education, Xi’an Jiaotong University, China
\and
{\tt\small rayn\_z@stu.xjtu.edu.cn, shibin@xjtu.edu.cn}
\and
{\tt\small jianfei.ruan@hotmail.com, terryptz@stu.xjtu.edu.cn, dong.bo@xjtu.edu.cn}
}

\begin{document}
\maketitle
\begin{abstract}
In noisy label learning, estimating noisy class posteriors plays a fundamental role for developing consistent classifiers, as it forms the basis for estimating clean class posteriors and the transition matrix. Existing methods typically learn noisy class posteriors by training a classification model with noisy labels. However, when labels are incorrect, these models may be misled to overemphasize the feature parts that do not reflect the instance characteristics, resulting in significant errors in estimating noisy class posteriors. To address this issue, this paper proposes to augment the supervised information with part-level labels, encouraging the model to focus on and integrate richer information from various parts. Specifically, our method first partitions features into distinct parts by cropping instances, yielding part-level labels associated with these various parts. Subsequently, we introduce a novel single-to-multiple transition matrix to model the relationship between the noisy and part-level labels, which incorporates part-level labels into a classifier-consistent framework. Utilizing this framework with part-level labels, we can learn the noisy class posteriors more precisely by guiding the model to integrate information from various parts, ultimately improving the classification performance. Our method is theoretically sound, while experiments show that it is empirically effective in synthetic and real-world noisy benchmarks.
\end{abstract}

\section{Introduction}
The increase in model capacity has enabled deep artificial neural networks to fit almost any data, which unfortunately makes them prone to memorizing even the mislabeled instances \cite{wu2021class2simi, yao2020dual}. Thus, to train an effective neural network, a large amount of accurately labeled data is required. However, obtaining accurate labeling in real-world tasks typically involves manual labeling, which is often time-consuming and costly \cite{xia2019anchor}. In contrast, massive amounts of noisy labels are readily available through web crawlers, questionnaires and crowdsourcing \cite{li2019dividemix}.

To reduce the negative impact of noisy labels, various heuristic strategies are applied to Noisy Label Learning (NLL), such as selecting reliable samples \cite{li2019dividemix, yao2020searching, cheng2020learning, karim2022unicon, wang2022scalable, xia2021sample, huang2022uncertainty} and correcting labels \cite{tanaka2018joint, li2022neighborhood, wu2022boundaryface}. Although these methods can train classifiers that perform well empirically, they do not guarantee a \emph{consistent classifier}, meaning that the classifiers learned from noise-labeled data will not necessarily converge to the optimal one learned from clean data \cite{yao2020dual, xia2019anchor}. 

To overcome this problem, various \emph{classifier-consistent} methods have been proposed, with the most successful methods employing loss correction procedures \cite{patrini2017making, hendrycks2018using, xia2019anchor, xia2020part, yao2020dual, li2021provably, xia2022extended, cheng2022instance, yang2022estimating}. The basic idea of these methods is that, the \emph{clean class posterior} $P(\boldsymbol{Y}|\boldsymbol{x})$ can be inferred from \emph{noisy class posterior} $P(\tilde{\boldsymbol{Y}}|\boldsymbol{x})$ and \emph{transition matrix} $T(\boldsymbol{x})$, from the equation $P(\tilde{\boldsymbol{Y}}|\boldsymbol{x}) = T(\boldsymbol{x})^\top P(\boldsymbol{Y}|\boldsymbol{x})$ \footnote{We define $P(\boldsymbol{Y}|\boldsymbol{x})=[P(Y = 1|X = \boldsymbol{x}), . . . , P(Y = c|X = \boldsymbol{x})]^\top$ and $T_{ij}(\boldsymbol{x})=P(\tilde{Y}=j|Y=i, X=\boldsymbol{x})$, where $c$ denotes the number of classes, $X$ and $Y$ represents the variable for clean labels and instances, respectively.}. The noisy class posteriors estimation plays a fundamental role in this process, directly affecting the calculation of clean class posteriors while also being critical for the accurate transition matrix estimation \cite{cheng2022class}.
\begin{figure*}[t]
  \centering
  \begin{minipage}[t]{\linewidth}
    \centering
    \subfloat[\label{fig:before}]{\includegraphics[width=\linewidth]{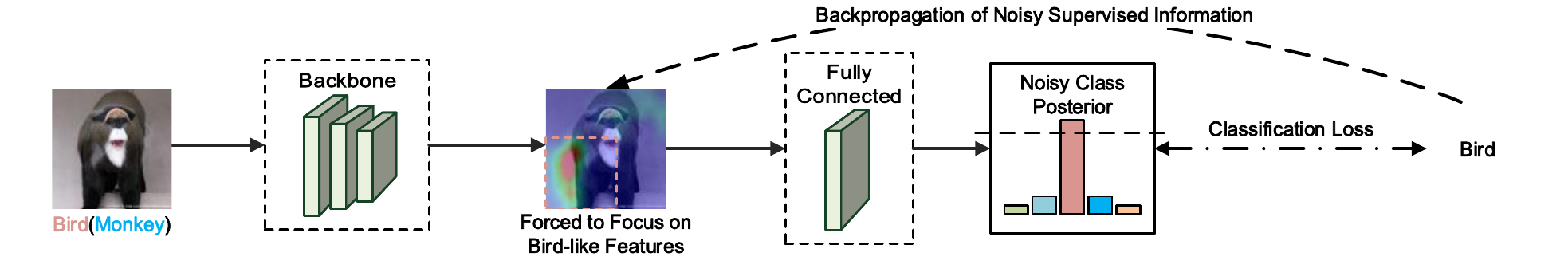}}
  \end{minipage}
  \begin{minipage}[t]{\linewidth}
    \centering
    \subfloat[\label{fig:method}]{\includegraphics[width=\linewidth]{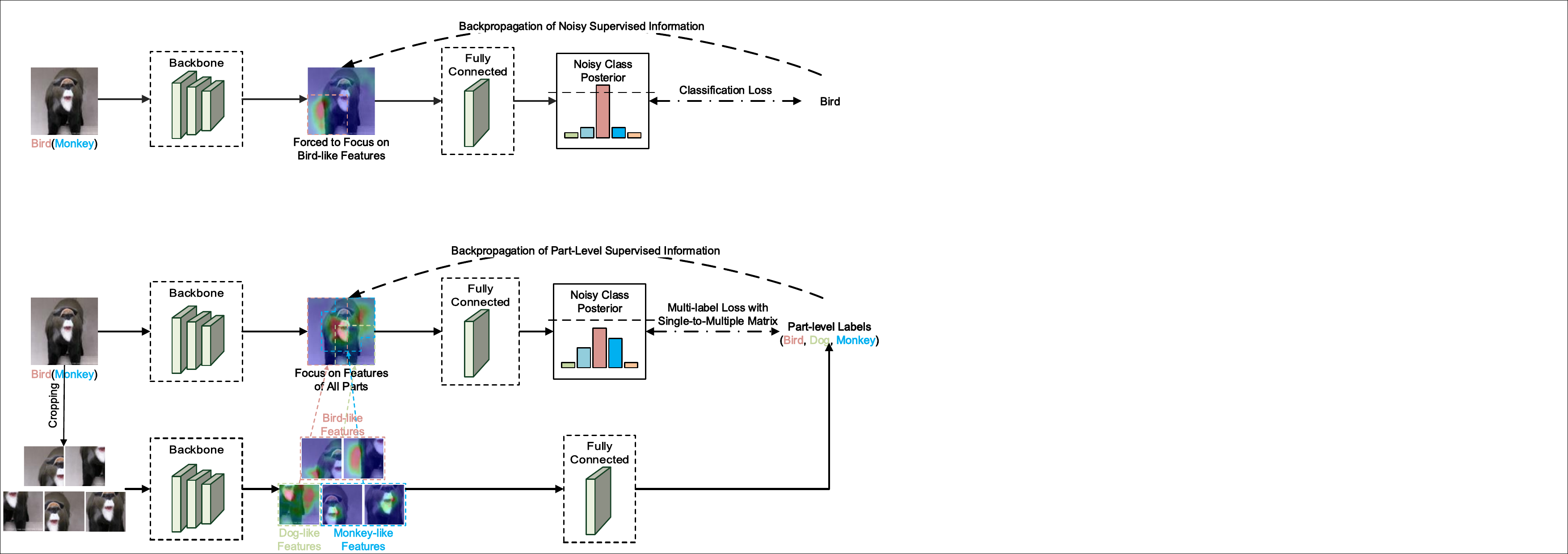}}
  \end{minipage}
  \caption{Illustration of overemphasis that arises when learning noisy class posterior with classification loss (Figure \ref{fig:before}), as well as the framework we proposed to alleviate this overemphasis (Figure \ref{fig:method}). Class activation maps are used to visualize feature importance for estimation, where the highlighted areas (with stronger red intensity) represent the focus regions of the model.} 
  \label{fig:all}
\end{figure*} 
Considering that the ground-truth noisy class posterior is typically unavailable, existing methods often learn it through the classification tasks supervised by noisy labels \cite{patrini2017making, hendrycks2018using, xia2019anchor, xia2020part, yao2020dual, li2021provably, xia2022extended, cheng2022instance, yang2022estimating}. In classification, labels will encourage the model to focus on label-related feature part to minimize the classification loss. However, when labels are incorrect, models will be misled into overemphasizing the erroneous parts that do not reflect the instance characteristics. As illustrated in Figure \ref{fig:before}, for an instance of a "feathered monkey" labeled as "bird", the noisy label "bird" will force the backbone network to focus on features related to birds, such as feathers, while ignoring other important features, such as monkey's facial features, to achieve a smaller loss through backpropagation. In such cases, the model's output tends to reflect the probability of 'birds' given "feathers" features rather than "birds" given this "feathered monkey" instance. This introduces significant errors in estimating the noisy class posteriors. To address this issue, a straightforward idea is to augment supervised information with part-related labels, thus encouraging the model to focus on disparate feature parts. This can enhance the model's ability to capture instance characteristics and aid in estimating noisy class posteriors. This idea is consistent with object recognition research, which reveals that feature parts are used by humans and machines to identify objects \cite{xia2020part}, drawing on evidence from psychological and physiological studies \cite{palmer1977hierarchical, wachsmuth1994recognition, logothetis1996visual}, as well as computational theories and learning algorithms \cite{biederman1987recognition, ullman1996high, dietterich1997solving, norouzi2014zero, hosseini2015deep, agarwal2004learning}.

Motivated by this idea, we propose a Part-Level Multi-labeling (PLM) method, which generates multiple part-level labels to guide the model's focus. As illustrated in Figure \ref{fig:method}, by augmenting the supervised information, we attempt to capture distinct features of various parts, thereby rectifying the model's excessive focus on specific misleading parts. Specifically, to generate part-level labels, we design a multi-labeling approach based on instance cropping. As displayed in the second row of Figure \ref{fig:method}, this approach partitions features into distinct parts by instance cropping, such that some of the parts do not contain excessively focused features. By means of this partitioning, we can employ a network trained on noisy data to assign labels to these parts, resulting in multiple part-level labels associated with various diverse focused regions. Subsequently, a single-to-multiple transition matrix is utilized to model the relationship between the single noisy label and the multiple part-level labels for each instance. With this matrix, we proposed a label joint training framework that incorporates both part-level and noisy supervised information into a classifier-consistent framework. As shown in the first row of Figure \ref{fig:method}, this framework guides the model to focus on more diverse regions, thereby obtaining representations that reflect instance characteristics, ultimately enhancing the learning of noisy class posteriors. Then, it can be used to estimate other critical metrics such as transition matrix or serve as a component of loss correction for NLL. We extensively evaluated our proposed method on synthetic datasets with instances-independent and -dependent noise, as well as on real-world datasets, offering empirical evidence of its efficacy in reducing errors in noisy class posterior estimation and improving performance on NLL tasks.

Our main contributions are summarized as follows:
\begin{itemize}
   \item We focus on a novel problem of estimating noisy class posteriors in noisy label learning, which forms the basis for building classifier-consistent algorithms.
   \item We are the first to note the misleading effect of incorrect labels on noisy class posteriors, where incorrect labels guide the model to overly focus on feature details that do not reflect instance characteristics. To counter this, we propose to incorporate part-related supervision by a consistent classifier, which guides the model to integrate information from various parts.

   
   
   \item Extensive experiments on a variety of synthetic and real-world noisy datasets have confirmed the effectiveness of proposed method. The method notably enhances the estimation of noisy class posteriors and can be integrated with various NLL methods that rely on noisy class posteriors to boost their classification performance.
\end{itemize}

\section{Related work}
The primary goal of NLL is to mitigate the impact of mislabeled data in the training, fostering more robust models. In this context, various data-centric \cite{li2019dividemix, yao2020searching, cheng2020learning, karim2022unicon, wang2022scalable, tanaka2018joint, li2022neighborhood, wu2022boundaryface, ren2018learning, han2018masking, li2020gradient, han2018co, nguyen2020self, yao2021instance, garg2023instance, li2022selective} and loss-centric \cite{zhang2018generalized, xu2019l_dmi, lyu2019curriculum, liu2020peer, patrini2017making, hendrycks2018using, xia2019anchor, xia2020part, yao2020dual, li2021provably, xia2022extended, cheng2022instance, yang2022estimating, li2022estimating} approaches have been proposed. The former involves data preprocessing to reduce the involvement of noisy data in training, while the latter constructs robust losses to diminish the strength of the supervised signals generated by noisy data.

The first category of methods often uses heuristic algorithms to reduce the side effects of error data, such as selecting reliable samples \cite{li2019dividemix, yao2020searching, cheng2020learningsieve, karim2022unicon, wang2022scalable}, correcting labels \cite{kremer2018robust, tanaka2018joint, li2022neighborhood, wu2022boundaryface}, reweighting samples \cite{ren2018learning}, smoothing labels \cite{lukasik2020does} and noise reduction \cite{wu2021class2simi}. However, the use of heuristic algorithms makes these methods prone to overfitting noisy data. To address this issue, the second category of methods aims to introduce robust loss functions to weaken the supervised signals from mislabeled data, avoiding the reliance on heuristic algorithms. The most prominent approach within this category is \emph{loss correction} methods \cite{patrini2017making, hendrycks2018using, xia2019anchor, xia2020part, yao2020dual, li2021provably, xia2022extended, cheng2022instance, yang2022estimating}. These methods typically use a transition matrix to correct the loss to guarantee that the trained classifier will converge to the optimal classifier learned from the clean data (i.e., statistically consistent classifier). The critical task of this method is to estimate the transition matrix and thus infer the clean class posterior from the noisy data. 

In these methods, the noisy class posterior is often utilized to estimate important parameters, such as calculating losses to obtain low-loss samples and estimating the transition probabilities to obtain a transition matrix, and even inferring clean class posteriors with a consistent classifier. Unfortunately, the estimation of these important metrics are all affected by the estimation errors of noisy class posteriors, which is considered a unresolved bottleneck of classification performance \cite{li2021provably, cheng2022class}. While some label processing methods \cite{guo2017calibration, lukasik2020does, wu2021class2simi} implicitly reduce overemphasis on specific parts by decreasing the model's fitting to noisy labels, they usually apply the same treatment to all labels, lacking an effective guidance to help the network integrate information from other parts. Hence, these methods struggle to address the issue where incorrect labels mislead the network to focus on features that do not reflect the instance characteristics.
\section{Method}
\subsection{Problem setting}
Let $D$ denote the joint probability distribution of a pair of random variables $(X, Y) \in \mathcal{X} \times \mathcal{C}$, where $X$ and $Y$ represent the random variables associated with instances and their corresponding clean labels, respectively. In this context, $\mathcal{X}$ denotes the instance space, while $\mathcal{C}=\{1,\dots,c\}$ represents the label space with $c$ denoting the number of classes. Given a labeled training dataset $\mathcal{D}=\{(\boldsymbol{x}_i,y_i)\}_{i=1}^n$ with size $n$, where each example $(\boldsymbol{x}_i, y_i)$ is drawn independently from the probability distribution $D$, the classification task aims to learn a classifier $f:\mathcal{X} \rightarrow \mathcal{C}$ that maps each instance $\boldsymbol{x}_i$ to its corresponding label $y_i$ based on the training data $\mathcal{D}$. However, obtaining samples from the distribution $D$ in real-world tasks presents a significant challenge, since the observed labels are often corrupted by noise. Let $\tilde Y$ be the random variable of noisy labels and $\tilde{D}$ be the distribution of a pair of random variables $(X, \tilde Y)$. The goal of NLL is to learn a robust classifier from noisy sample set $\tilde{\mathcal{D}} = \{(\boldsymbol{x}_i,\tilde{y_i})\}_{i=1}^n$ independently drawn from $\tilde{D}$, which can assign the clean labels for the instances.
\subsection{Preliminary: NLL with consistent classifiers} \label{sec:preliminary}
For building a consistent classifier, the mainstream methods using a transition matrix, which can relate the random variables of $\tilde{Y}$ and $Y$ \cite{cheng2020learning, xia2019anchor,cheng2022instance}, to infer the clean class posteriors from noisy class posteriors. Specifically, the noisy class posterior vector $P(\tilde{\boldsymbol{Y}}|\boldsymbol{x})=[P(\tilde{Y}=1|\boldsymbol{x}),\dots,P(\tilde{Y}=c|\boldsymbol{x})]^{\top}$ can be transformed into the multiplication of the clean class posterior vector $P(\boldsymbol{Y}|\boldsymbol{x})=[P(Y=1|\boldsymbol{x}),\dots,P(Y=c|\boldsymbol{x})]^{\top}$ and the noise transition matrix $T(\boldsymbol{x})\in \mathbb{R}^{c \times c}$. Here, $T(\boldsymbol{x})$ denotes the transition matrix of instance $\boldsymbol{x}$, and the element of the $i$-th row and $j$-th column of $T(\boldsymbol{x})$ is defined by $T_{ij}(\boldsymbol{x})=P(\tilde{Y}=j|Y=i,\boldsymbol{x})$. Specifically, according to \eq{eq:labeltrans}, there is $P(\tilde{\boldsymbol{Y}}|\boldsymbol{x})=T(\boldsymbol{x})^\top P(\boldsymbol{Y}|\boldsymbol{x})$. It means that the NLL task can be resolved by estimating the transition matrix $T(\boldsymbol{x})$ and posterior $P(\tilde{Y}|\boldsymbol{x})$ \cite{liu2015classification,xia2019anchor,xia2022extended}. 
\begin{equation} \label{eq:labeltrans}
  \begin{aligned}
    P(\tilde{Y}=j|\boldsymbol{x})&=\sum_{i=1}^{k}P(\tilde{Y}=j|Y=i,\boldsymbol{x})P(Y=i|\boldsymbol{x})\\
    &=\sum_{i=1}^{k}T_{ij}(\boldsymbol{x})P(Y=i|\boldsymbol{x}).
  \end{aligned}
\end{equation}

Thus, in the context of constructing a consistent classifier with a transition matrix, the accuracy of clean class posteriors directly relies on the estimation of noisy class posteriors. Besides, the estimation of the transition matrix, in many mainstream methods, also depends on noisy class posteriors \cite{cheng2022class}. Therefore, the estimation of noisy class posteriors is crucial in building classifier-consistent algorithms. To reduce errors in estimating noisy class posteriors, this paper introduces a training framework with a single-to-multiple transition matrix. It can serve as a component in loss correction methods to build a classifier-consistent NLL algorithm. Section \ref{sec:framework} provides a detailed discussion on the label joint training framework.

\subsection{Part-level multi-labeling}
This section provides a detailed explanation of the process of generating part-level labels through instance cropping, as illustrated in the second row of Figure \ref{fig:method}. Specifically, for each instance, we obtain a set of cropped parts by applying $K$ times crop, each capturing different features. Then, a noisy classifier trained on the raw noisy data is employed to assign labels to these parts, resulting in part-level labels that reflect diverse features. Since many parts do not contain the overemphasized features after cropping, the network can attend to other features when assigning labels to them, rather than being restricted by the features that are strongly correlated with the noisy labels. This leads to more informative labels that better reflect the comprehensive information of the instance.

Formally, for an instance $\boldsymbol{x}_i$, the goal of instance cropping is to generate a set of sub-instances $\mathcal{S}_i=\{\boldsymbol{s}_{i1},\dots,\boldsymbol{s}_{iK}\}^\top$, where $\boldsymbol{s}_{ij}$ signifies the $j$-th part of $\boldsymbol{x}_i$. Given a cropping number $K$, the results of instance cropping are influenced by the chosen cropping operation, which can be guided by different criteria, such as user-specified location, saliency, or attention maps. In the experiments conducted in this paper, we use image data as an example and crop five equally-sized parts uniformly from the four corners and the central position. After the instance cropping, we employ a labeling classifier network $f^l$ trained from noisy data to label each part. Part-level labels are a form of multi-label, and thus can be decomposed into $c$ independent binary labels, each corresponding to one of the possible labels in the label space. Thus, the part-level labels of instance $\boldsymbol{x}_i$ can be represented as a vector $\boldsymbol{y}^\prime_i=[y^\prime_{i1},\cdots,y^\prime_{ic}]$, where $y^\prime_{ij}$ denotes the part-level label associated with the $j$-th class. The value of $y^\prime_{ij}$ can be calculated as follows:
\begin{equation}
  y^\prime_{ij}=
        \begin{cases}
          1 & \exists \boldsymbol{s} \in \mathcal{S}_i, f^l(\boldsymbol{s})=j \\
          0 & \forall \boldsymbol{s} \in \mathcal{S}_i, f^l(\boldsymbol{s})\neq j
        \end{cases}.
\end{equation}
Through the labeling of distinct parts, the model is compelled to focus on and label each part independently. Then, these labels can guide the model to focus on and integrate the corresponding parts to avoid misleading.
\subsection{Label joint training framework} \label{sec:framework}
To leverage the part-level labels for estimating the noise class posterior, we propose a label joint training framework. This framework uses noisy labels and part-level labels jointly to estimate the noise class posterior, based on a single-to-multiple transition matrix that models how a single noisy label relates to multiple part-level labels.

As discussed in the previous section, the part-level labels of an instance can be interpreted as a multi-label. Therefore, the random variable associated with the part-level labels can be represented by a joint distribution $(Y_{1}^{\prime},\cdots,Y_{c}^{\prime})$, where $Y_{j}^{\prime} \in \{0, 1\}$ represents the random variable associated with the $j$-th label component of the part-level labels. Then, similar to the discussion in Section \ref{sec:preliminary}, the part-level labels can be linked to the noisy labels as follows:
\begin{equation}
  P(Y^\prime_j=1|X=\boldsymbol{x})=\sum_{i=1}^{c}U_{ij}(\boldsymbol{x})P(\tilde{Y}=i|X=\boldsymbol{x}),
  \label{eq:noise2part-level}
\end{equation}
where $U_{ij}(\boldsymbol{x})=P(Y^\prime_j=1|\tilde{Y}=i,X=\boldsymbol{x})$ is the $ij$-th entry of matrix $U(\boldsymbol{x}) \in \mathbb{R}^{c \times c}$, and $U(\boldsymbol{x})$ denotes the transition matrix of instance $\boldsymbol{x}$. In this way, we implemented the joint integration of part-level and noisy labels within a classifier-consistent framework.

\smallskip\noindent\textbf{Joint training with single-to-multiple matrix.} For estimating single-to-multiple transition matrix and noisy class posterior, we construct a single-to-multiple transition matrix estimation network $g^u: \mathcal{X}\rightarrow \mathbb{R}^{c \times c}$ and a noisy class posterior estimation network $g^e: \mathcal{X}\rightarrow \mathbb{R}^c$. We perform a joint training of $g^u$ and $g^e$ by minimizing the following loss function:
\begin{equation}
  \mathcal{L}(\boldsymbol{x}, \tilde{y}, \boldsymbol{y}^\prime)=\frac{1}{2}(\ell_c(g^e(\boldsymbol{x}),\tilde{y})+\ell_m(g^p(\boldsymbol{x}), \boldsymbol{y}^\prime)), \label{eq:pre_target}
\end{equation}
where network $g^p(\boldsymbol{x})=g^u(\boldsymbol{x})g^e(\boldsymbol{x})$, $\ell_c$ and $\ell_m$ represent a classification loss (e.g., the cross entropy loss) and a multi-label classification loss (e.g., the binary cross-entropy loss), respectively. By minimizing the loss function $\ell_c$, the output of network $g^e$ is adjusted to fit the noisy class posterior $\hat{P}(\tilde{\boldsymbol{Y}}|\boldsymbol{x})$. Furthermore, by minimizing $\ell_m$, $g^u(\boldsymbol{x})g^e(\boldsymbol{x})$ will output part-level multi-labels. Therefore, based on Eq. \eqref{eq:noise2part-level}, $g^u(\boldsymbol{x})$ is compelled to model the single-to-multiple transition matrix. As the estimation of $g^u$ becomes more accurate, the noisy class posterior estimation of $g^e$ also improves by minimizing $\ell_m$. By employing the joint training approach, we ensure that the noisy class posterior estimation benefits from the supervision of noisy labels and partial-level labels as Eq. \eqref{eq:noise2part-level}. The richer part contextual supervision prevents the overemphasis by $g^e$ on specific part, thus avoiding overconfidence. Here, we introduce a novel matrix, and its identifiability is discussed in Appendix B.

\smallskip\noindent\textbf{Classification with consistent classifier.} Beyond enhancing the efficiency of noise class posterior estimation to improve the estimation precision of other crucial metrics, our method can also function as a classifier-consistent component to be incorporated into a consistent classifier. From \eq{eq:labeltrans} and \eq{eq:noise2part-level}, we can formulate the relationship between the part-level labels and the clean labels as follows:
\begin{equation}
  \begin{aligned}
    P(Y^\prime_j=1|\boldsymbol{x})&=\sum_{i=1}^{c}U_{ij}(\boldsymbol{x})P(\tilde{Y}=i|\boldsymbol{x}) \\
    &=\sum_{i=1}^{c}U_{ij}(\boldsymbol{x})\sum_{k=1}^{c}T_{ki}(\boldsymbol{x})P(Y=k|\boldsymbol{x}).
  \end{aligned}
  \label{eq:doubletrans}
\end{equation}
The estimation of the transition matrix $T_{ki}(\boldsymbol{x})$ can be undertaken using existing methods. Subsequently, our framework can be exploited to amplify the effectiveness of NLL. The details of classifier training can be found in Appendix A.
\section{Experiments}
\subsection{Experiment setup} \label{sec:setup}
\smallskip\noindent\textbf{Datasets.} We evaluated the performance of our proposed model using both synthetic and real-world datasets. The synthetic datasets included MNIST \cite{lecun1998mnist}, CIFAR-10 \cite{krizhevsky2009learning} and CIFAR-100 \cite{krizhevsky2009learning}, which we generated by applying three types of noise: symmetry flipping \cite{patrini2017making}, pair flipping \cite{han2018co} and instance-dependent label noise (IDN) \cite{xia2020part}. The MNIST dataset consists of 60,000 training images and 10,000 test images, representing handwritten digits from 0 to 9. The CIFAR-10 and CIFAR-100 datasets consist of 50,000 training images and 10,000 test images, representing 10 and 100 object classes, respectively. We also used Clothing1M \cite{xiao2015learning}, a large-scale dataset of 1M real-world clothing images with noisy labels for training and 10k images with clean labels for testing. The dataset also provides 50k and 14k clean labels for training and validation, but we did not use them following the settings of VolMinNet \cite{li2021provably}. For model selection, we randomly sampled 10\% of the noisy training data as validation sets from each dataset.

\smallskip\noindent\textbf{Backbone and implementation details.} For MNIST, we use a Lenet \cite{lecun1998gradient} as backbone network. The Stochastic Gradient Descent (SGD) optimizer is used to train the network, with an initial learning rate $5\times 10^{-2}$, weight decay $10^{-4}$ and momentum 0.9. For CIFAR-10, we use a ResNet-18 \cite{he2016identity} as backbone network. The SGD optimizer is used to train the network, with an initial learning rate $10^{-2}$, weight decay $10^{-2}$, momentum 0.9. We use the cosine learning rate decay strategy for MNIST and CIFAR-10. For CIFAR-100, we use a ResNet-32 \cite{he2016identity} as backbone network. SGD optimizer is used to train the network, with an initial learning rate $5 \times 10^{-2}$, weight decay $10^{-3}$, momentum 0.9. The learning rate is divided by 100 after the 40-th epoch. We apply random horizontal flips and random crops of size $32 \times 32$ pixels after adding a 4-pixel padding on each side for both CIFAR-10 and CIFAR-100 datasets. For all synthetic datasets, the labeling network and estimation network are both trained for 50 epochs with batch size 128 with same parameters. For clothing1M, we use a ResNet-50 \cite{he2016identity} pretrained on ImageNet as backbone network. SGD optimizer is used to train the network, with an initial learning rate $10^{-2}$, weight decay $10^{-3}$, momentum 0.9. The learning rate is divided by 10 after the 10-th epoch and the 20-th epoch. We resize the image to $256 \times 256$ pixels and apply random crops of size $224 \times 224$. The labeling network and estimation network are both trained 30 epochs with batch size 128, and we undersample the noisy labels for each class in every epoch to balance them like VolMinNet\cite{li2021provably}. To generate sub-instances, we select five parts uniformly at the four corners and one central position. The part sizes are $22 \times 22$ pixels for MNIST, $25 \times 25$ pixels for CIFAR-10 and CIFAR-100, and $179 \times 179$ pixels for Clothing1M. Additionally, we discussed and conducted experiments on various instance cropping approach in Appendix D. The code are available at https://github.com/RyanZhaoIc/PLM.
\subsection{Noisy class posterior estimation}
\begin{figure*}[htb]
  \centering
  \includegraphics[width=0.95\linewidth]{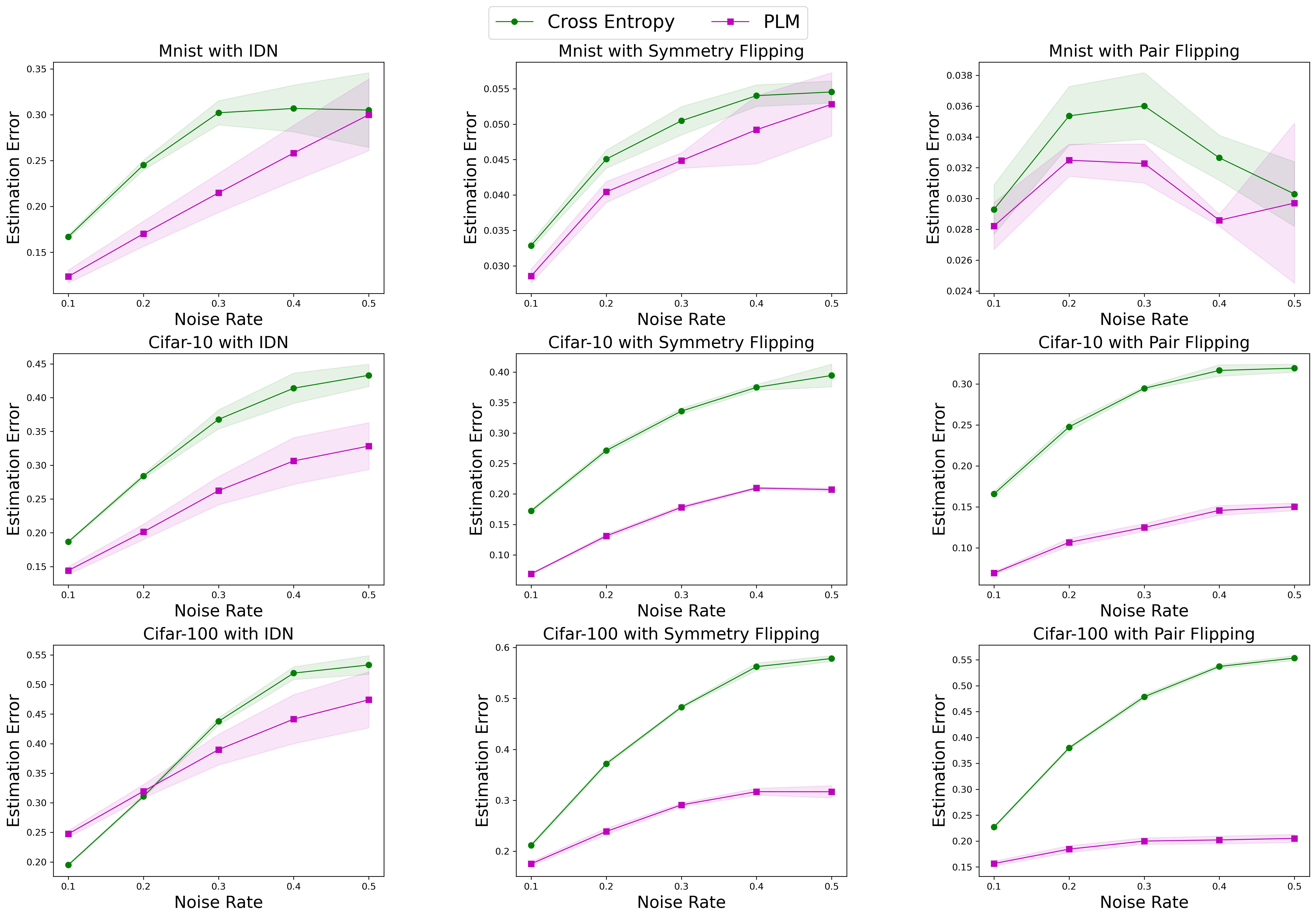}
  \caption{Mean estimation errors of noise class posterior for 5 trials on Mnist, CIFAR-10 and CIFAR-100. The error bars of standard deviation are shaded in each plot. The lower the better.} 
  \label{fig:errors}
  \vspace{-10pt}
\end{figure*} 
We compared PLM with the CE approach in terms of the estimation error of noisy class posteriors across diverse datasets and noise rates. This comparison is pertinent as the previous loss correction methods \cite{patrini2017making, xia2019anchor, yao2020dual,li2021provably,xia2022extended} predominantly utilize the CE loss to estimate the posterior probability of noisy labels. For parity in comparison, we implemented both methods using the same backbone network as Section \ref{sec:setup}. The estimation error is calculated as the $l_2$-distance between the ground-truth and the estimated noisy class posteriors of on the pseudo-anchor dataset, where the ground-truth originates from the noise generation process. Specifically, we first train a model with clean labels to select high-confidence samples as pseudo-anchor dataset. Then we assume that a pseudo-anchor sample $(\boldsymbol{x},i)$ satisfies $P(Y=i|X=\boldsymbol{x})=1$, thereby the ground-truth noisy class posterior of $(\boldsymbol{x})$ can be calculated as $P(\tilde{\boldsymbol{Y}}|X=\boldsymbol{x})=P(\tilde{\boldsymbol{Y}}|Y=i,X=\boldsymbol{x})$. Thus, on the synthetic datasets, noise class posteriors can be obtained through the defined noise flipping probability $P(\tilde{\boldsymbol{Y}}|Y=i,X=\boldsymbol{x})$. This setting are similar to the Dual-T \cite{yao2020dual}. 

We conducted these experiments across noise rates of $[0.1, 0.2, 0.3, 0.4, 0.5]$. As depicted in Figure \ref{fig:all}, our method and the comparison methods demonstrated similar trends in the estimation error of noisy class posterior probabilities. However, the estimation error of PLM is consistently smaller than that of the CE at different noise rates, especially at higher ones. The results indicate that our method can provide estimations that are closer to the true noisy class posterior probabilities. The visualization in Appendix E can also serve as evidence for this result.
\subsection{Comparison for classification}
\begin{table*}[t]
  \centering
  \caption{The average classification accuracy and standard deviation (expressed in percentage) across five trials on the MNIST, CIFAR-10, and CIFAR-100 datasets under various synthetic noisy label settings. The best classification accuracy is indicated in \textbf{bold}.}
  \vspace{-2pt}
  \label{tab:accuracy}
  \begin{tabular}{cccccccc}
      \toprule
      & \multicolumn{2}{c}{MNIST} & \multicolumn{2}{c}{CIFAR-10} & \multicolumn{2}{c}{CIFAR-100}\\ 
      & Sym-20\% & Sym-50\% & Sym-20\% & Sym-50\% & Sym-20\% & Sym-50\% &  \\ 
      \midrule
      Decoupling & 97.58 $\pm$ 0.16 & 95.66 $\pm$ 0.17 & 87.11 $\pm$ 0.24 & 77.62 $\pm$ 0.30 & 69.71 $\pm$ 0.34 & 48.30 $\pm$ 0.68 &  \\ 
      Co-teaching & 96.35 $\pm$ 0.10 & 92.01 $\pm$ 0.24 & 78.09 $\pm$ 0.91 & 67.16 $\pm$ 0.32 & 44.89 $\pm$ 0.63 & 32.19 $\pm$ 0.90 &  \\ 
      T-Revision & 98.86 $\pm$ 0.04 & 98.41 $\pm$ 0.08 & 89.41 $\pm$ 0.18 & 83.68 $\pm$ 1.07 & 61.14 $\pm$ 0.69 & 40.01 $\pm$ 0.90 &  \\ 
      Dual-T & 98.82 $\pm$ 0.17 & 98.24 $\pm$ 0.14 & 89.79 $\pm$ 0.40 & 77.97 $\pm$ 1.91 & 65.61 $\pm$ 0.39 & 50.07 $\pm$ 1.56 &  \\ 
      DMI & 98.87 $\pm$ 0.09 & 96.96 $\pm$ 0.68 & 84.65 $\pm$ 0.83 & 70.62 $\pm$ 3.73 & 51.89 $\pm$ 2.24 & 32.02 $\pm$ 1.36 &  \\ 
      VolMinNet & 98.82 $\pm$ 0.12 & 98.13 $\pm$ 0.15 & 90.19 $\pm$ 0.17 & 84.09 $\pm$ 0.68 & 67.70 $\pm$ 0.99 & 57.99 $\pm$ 0.40 &  \\ 
      Class2Simi & 99.11 $\pm$ 0.06 & 98.33 $\pm$ 0.26 & 84.92 $\pm$ 0.50 & 70.02 $\pm$ 3.27 & 49.78 $\pm$ 1.05 & 32.02 $\pm$ 1.33 &  \\ 
      Forward & 98.47 $\pm$ 0.19 & 97.78 $\pm$ 0.20 & 85.49 $\pm$ 1.12 & 74.33 $\pm$ 1.31 & 58.70 $\pm$ 0.73 & 39.82 $\pm$ 2.15 &  \\ 
      PLM & \textbf{99.32 $\pm$ 0.02} & \textbf{98.96 $\pm$ 0.07} & \textbf{91.10 $\pm$ 0.24} & \textbf{85.08 $\pm$ 0.16} & \textbf{69.54 $\pm$ 0.23} & \textbf{60.44 $\pm$ 0.28} &  \\ 
      \midrule
      \midrule
      & Pair-20\% & Pair-45\% & Pair-20\% & Pair-45\% & Pair-20\% & Pair-45\% &  \\ 
      \midrule
      Decoupling & 97.37 $\pm$ 0.16 & 84.01 $\pm$ 1.90 & 88.93 $\pm$ 0.38 & 70.55 $\pm$ 2.29 & 66.65 $\pm$ 0.27 & 43.82 $\pm$ 1.75 &  \\ 
      Co-teaching & 95.38 $\pm$ 0.29 & 89.17 $\pm$ 0.27 & 79.18 $\pm$ 0.44 & 68.58 $\pm$ 0.68 & 45.78 $\pm$ 1.02 & 28.98 $\pm$ 0.42 &  \\ 
      T-Revision & 99.07 $\pm$ 0.10 & 96.42 $\pm$ 4.41 & 91.16 $\pm$ 0.16 & 81.88 $\pm$ 13.30 & 60.56 $\pm$ 0.66 & 49.44 $\pm$ 1.64 &  \\ 
      Dual-T & 98.86 $\pm$ 0.09 & 92.47 $\pm$ 5.91 & 89.59 $\pm$ 1.34 & 74.39 $\pm$ 4.30 & 71.08 $\pm$ 0.27 & 52.25 $\pm$ 3.28 &  \\ 
      DMI & 98.81 $\pm$ 0.13 & 96.45 $\pm$ 0.54 & 87.66 $\pm$ 0.69 & 80.30 $\pm$ 3.38 & 43.63 $\pm$ 0.86 & 32.22 $\pm$ 3.42 &  \\ 
      VolMinNet & 99.11 $\pm$ 0.08 & 99.10 $\pm$ 0.08 & 91.55 $\pm$ 0.13 & 86.12 $\pm$ 1.26 & 71.65 $\pm$ 0.62 & 61.21 $\pm$ 2.98 &  \\ 
      Class2Simi & 98.97 $\pm$ 0.12 & 98.33 $\pm$ 0.26 & 87.27 $\pm$ 0.31 & 75.47 $\pm$ 6.09 & 53.39 $\pm$ 0.95 & 36.04 $\pm$ 2.31 &  \\ 
      Forward & 98.82 $\pm$ 0.15 & 91.02 $\pm$ 10.81 & 89.33 $\pm$ 1.09 & 69.99 $\pm$ 9.79 & 60.12 $\pm$ 0.33 & 37.99 $\pm$ 0.37 &  \\ 
      PLM & \textbf{99.40 $\pm$ 0.03} & \textbf{99.39 $\pm$ 0.05} & \textbf{92.84 $\pm$ 0.17} & \textbf{91.80 $\pm$ 0.63} & \textbf{72.67 $\pm$ 0.32} & \textbf{61.90 $\pm$ 1.82} &  \\ 
      \bottomrule
  \end{tabular}
\end{table*}
 
\begin{table*}[htb]
  \caption{The classification accuracy (expressed in percentage) on the Clothing1M dataset. Comparative baseline method results are quoted from PTD \cite{xia2020part} and VolMinNet \cite{li2021provably} since we employed the same experimental setup. The best classification accuracy is indicated in \textbf{bold}.}
  \vspace{-2pt}
  \label{tab:clothing1M}
  \centering
  \setlength{\tabcolsep}{1.5mm}{
  \begin{tabular}{cccccccc}
    \toprule
    \makebox[0.1\textwidth][c]{Co-teaching} & \makebox[0.1\textwidth][c]{Dual-T} & \makebox[0.1\textwidth][c]{T-Revision} & \makebox[0.1\textwidth][c]{DMI} & \makebox[0.1\textwidth][c]{VolMinNet} & \makebox[0.1\textwidth][c]{PTD} & \makebox[0.1\textwidth][c]{Forward} & \makebox[0.1\textwidth][c]{PLM}\\ 
    \midrule
    56.79 & 70.97 & 60.15 & 70.12 & 71.67 & 72.42 & 69.91 & \textbf{73.30}\\ 
    \bottomrule
  \end{tabular}
  }
  \vspace{-10pt}
\end{table*}
\begin{table}[t]
  \caption{The average classification accuracy and standard deviation (expressed in percentage) across five trials on the CIFAR-10 dataset with IDN. The results of the compared baseline methods are quoted from MEIDTM \cite{cheng2022instance} since we used the same experimental setup. The best classification accuracy is indicated in \textbf{bold}.}
  \vspace{-2pt}
  \label{tab:idn}
  \centering
  \resizebox{0.99\columnwidth}{!}{
  \begin{tabular}{ccccc}
    \toprule
    & IDN-20\% & IDN-30\% & IDN-40\% & IDN-50\% \\
    \midrule
    Co-teaching & 88.43 $\pm$ 0.08 & 86.40 $\pm$ 0.41 & 80.85 $\pm$ 0.97 & 62.63 $\pm$ 1.51 \\
    DMI & 89.99 $\pm$ 0.15 & 86.87 $\pm$ 0.34 & 80.74 $\pm$ 0.44 & 63.92 $\pm$ 3.92  \\
    PTD & 89.33 $\pm$ 0.70 & 85.33 $\pm$ 1.86 & 80.59 $\pm$ 0.41 & 64.58 $\pm$ 2.86  \\
    TMDNN & 88.14 $\pm$ 0.66 & 84.55 $\pm$ 0.48 & 79.71 $\pm$ 0.95 & 63.33 $\pm$ 2.75  \\
    MEIDTM & 91.38 $\pm$ 0.34 & 87.68 $\pm$ 0.26 & 82.63 $\pm$ 0.24 & 72.17 $\pm$ 1.51  \\
    Forward & 89.62 $\pm$ 0.14 & 86.93 $\pm$ 0.15 & 80.29 $\pm$ 0.27 & 65.91 $\pm$ 1.22  \\
    PLM & \textbf{91.41 $\pm$ 0.17} & \textbf{88.60 $\pm$ 0.53} & \textbf{83.98 $\pm$ 2.53} & \textbf{76.87 $\pm$ 1.59}  \\
    \bottomrule
  \end{tabular}
  }
  \vspace{-10pt}
\end{table}
\begin{table*}[htb]
  \caption{The average classification accuracy and standard deviation (expressed in percentage) across five trials on the CIFAR-10 and CIFAR-100 datasets.The better classification accuracy is indicated in \textbf{bold}.}
  \vspace{-2pt}
  \label{tab:accuracy-10}
  \centering
  \footnotesize
  \begin{tabular}{ccccccccc}
      \toprule
      &\multicolumn{4}{c}{CIFAR-10} & \multicolumn{4}{c}{CIFAR-100}\\
      & Sym-20\% & Sym-50\% & Pair-20\% & Pair-45\% & Sym-20\% & Sym-50\% & Pair-20\% & Pair-45\% \\ 
      \midrule
      Forward & 85.62 $\pm$ 0.58 & 73.78 $\pm$ 0.99 & 89.20 $\pm$ 1.56 & 66.20 $\pm$ 10.05 & 58.70 $\pm$ 0.73 & 39.82 $\pm$ 2.15 & 60.12 $\pm$ 0.33 & 37.99 $\pm$ 0.37 \\ 
      PLM-F & \textbf{89.45 $\pm$ 0.59} & \textbf{81.94 $\pm$ 0.62} & \textbf{91.29 $\pm$ 0.18} & \textbf{74.90 $\pm$ 1.37} & \textbf{68.19 $\pm$ 0.92} & \textbf{58.59 $\pm$ 0.92} & \textbf{70.94 $\pm$ 1.16} & \textbf{55.47 $\pm$ 1.27} \\  
      \midrule
      Dual-T & 89.79 $\pm$ 0.40 & 77.97 $\pm$ 1.91 & 89.59 $\pm$ 1.34 & 74.39 $\pm$ 4.30 & 65.62 $\pm$ 0.39 & 50.07 $\pm$ 1.56 & 71.08 $\pm$ 0.27 & 52.25 $\pm$ 3.28 \\ 
      PLM-D & \textbf{90.95 $\pm$ 0.28} & \textbf{85.38 $\pm$ 0.49} & \textbf{93.15 $\pm$ 0.26} & \textbf{91.10 $\pm$ 2.17} & \textbf{69.44 $\pm$ 0.27} & \textbf{60.84 $\pm$ 0.38} & \textbf{71.90 $\pm$ 0.56} & \textbf{68.38 $\pm$ 1.11} \\ 
      \midrule 
      T-Revision & 88.68 $\pm$ 0.54 & 82.59 $\pm$ 1.82 & 91.10 $\pm$ 0.21 & 72.65 $\pm$ 16.11 & 61.14 $\pm$ 0.69 & 40.01 $\pm$ 0.90 & 60.56 $\pm$ 0.66 & 49.44 $\pm$ 1.64 \\ 
      PLM-R & \textbf{91.33 $\pm$ 0.47} & \textbf{85.22 $\pm$ 0.54} & \textbf{92.69 $\pm$ 0.23} & \textbf{91.33 $\pm$ 0.60} & \textbf{67.52 $\pm$ 1.19} & \textbf{50.81 $\pm$ 3.31} & \textbf{70.58 $\pm$ 1.13} & \textbf{56.85 $\pm$ 1.73} \\ 
      \midrule 
      VolMinNet & 90.19 $\pm$ 0.17 & 84.09 $\pm$ 0.68 & 91.55 $\pm$ 0.13 & 86.12 $\pm$ 1.26 & 67.70 $\pm$ 0.99 & 57.99 $\pm$ 0.40 & 71.65 $\pm$ 0.62 & 61.21 $\pm$ 2.98 \\ 
      PLM-V & \textbf{91.75 $\pm$ 0.19} & \textbf{84.10 $\pm$ 0.55} & \textbf{93.40 $\pm$ 0.23} & \textbf{86.91 $\pm$ 1.02} & \textbf{70.95 $\pm$ 0.49} & \textbf{62.37 $\pm$ 0.29} & \textbf{74.55 $\pm$ 0.12} & \textbf{64.05 $\pm$ 0.53} \\
      \bottomrule
  \end{tabular}
  \vspace{-10pt}
\end{table*}
To compare the accuracy of the NLL, we first estimate the noise transition matrix using the same approach as Forward \cite{patrini2017making}. Subsequently, we train a single-to-multiple transition matrix estimation network using the method proposed in this paper. Afterward, we fix the transition matrix estimation network and train the classification network with a method like Forward, following the label transition relationship in \eq{eq:doubletrans} and the training objective in \eq{eq:pre_target}. Here, we use the average of the 10 estimated anchors’ predicted probabilities as a row in the transition matrix, and train the classification network with the same parameters as the estimation network in Section \ref{sec:setup}. We compare the classification accuracy of PLM with the following baseline methods: (1) Decoupling \cite{malach2017decoupling}, which trains two predictors and updates them only on examples that they disagree on; (2) Co-teaching \cite{han2018co}, which trains two deep networks and exchanges the examples with small loss for network updating; (3) Forward \cite{patrini2017making}, which corrects the training loss by the class flipping transition matrix; (4) T-Revision \cite{xia2019anchor}, which estimates transition matrix with a slack variable to correct training loss; (5) Dual-T \cite{yao2020dual}, which estimates transition matrix by factorizing the original transition matrix with an intermediate class; (6) DMI \cite{xu2019l_dmi}, which handles label noise by using information-theoretic loss based on the determinant of the joint distribution matrix; (7) VolMinNet \cite{li2021provably}, which estimates the matrix that incorporates the minimum volume constraint into the label-noise learning; (8) Class2Simi \cite{wu2021class2simi}, which transforms noisy class labels into noisy similarity labels to reduce the noise rate. For a fair comparison, we implement all methods with the same backbone network and default parameters on each dataset by the PyTorch and run all experiments on the NVIDIA RTX 3090 GPUs. It should be noted that we did not compare our method with some state-of-the-art methods, such as DivideMix\cite{li2019dividemix} and NCR\cite{iscen2022learning}. These methods incorporate multiple advanced methods (e.g., semi-supervised learning, contrastive learning and complex data augmentation, etc.) to improve their performance under noisy labels, while PLM is solely focused on reducing the error of estimating noisy class posterior to benefit the NLL, negating the need for additional techniques. Thus, a direct comparison would not present a fair assessment.

In Table \ref{tab:accuracy}, we report the classification accuracy of PLM and all baseline methods, under varying noise generation methods and rates. Here, we use Sym-$\epsilon$ and Pair-$\epsilon$ to denote the symmetry flipping and pair flipping methods with noise rates of $\epsilon$, respectively. Compared to other baseline methods, PLM shows superior classification accuracy. We executed five experiments on synthetic noisy datasets with distinct random seeds, introducing a higher degree of label randomness. Despite varying data generations, our method demonstrates robust performance. Our method outperformed the comparative methods in all metrics. It is worth noting that although we did not combine PLM with the latest loss correction methods in this study, the PLM method still achieved competitive performance and significantly improved the classification performance of the model under high noise rates. At the same time, our method only exhibits a linear increase in computational time, and we provide relevant discussions in Appendix C. The results indicate that PLM can serve as an effective auxiliary component to improve the robustness of the model to label noise. 

Additionally, we report the results on the Clothing1M dataset in Table \ref{tab:clothing1M}, utilizing the same dataset settings as VolMinNet. Our proposed method demonstrates superior performance over the compared baseline methods on the Clothing1M dataset, reflecting a 3.91\% enhancement relative to the Forward method. Additionally, we reported the performance on the real-world Animal-10N \cite{song2019selfie} dataset in Appendix H. These results of real-world datasets can be attributed to the PLM approach, which models instance information at the part-level, thereby bolstering the model's robustness against instance-dependent noise.
To further demonstrate this advantage, we conducted experiments in IDN synthetic noise in Table \ref{tab:idn}. Here, we adopted Forward's approach to estimate the noise transition matrix to evaluate how PLM improves the robustness to IDN using this method, and employ a backbone network in alignment with the experimental and network settings of PTD \cite{cheng2022instance}. We compare with the following baseline methods: (1) Co-teaching \cite{han2018co}; (2) DMI \cite{xu2019l_dmi}; (3) Forward \cite{patrini2017making}; (4) PTD \cite{xia2020part}, which uses a part-based noise transition matrix estimation technique; (5) TMDNN \cite{yang2022estimating}, which uses deep neural networks to estimate the transition matrix by exploiting Bayes optimal labels; (6) MEIDTM \cite{cheng2022instance}, which estimates IDN transition matrix by formulating a manifold assumption. Among them, PTD, TMDNN, and MEIDTM are methods that are designed for the IDN modeling. The results demonstrates that PLM achieves competitive performance without utilizing IDN modeling techniques. 

\smallskip\noindent\textbf{Learning with different noise transition matrices.} \label{sec:matrix} To assess the efficacy of our proposed method in combination with various noise transition matrix estimation techniques, we present classification experiment results on the CIFAR-10 and CIFAR-100 datasets in Table \ref{tab:accuracy-10}. Specifically, we merge PLM with the matrix estimation techniques outlined in Forward \cite{patrini2017making}, Dual-T \cite{yao2020dual}, T-Revision \cite{xia2019anchor}, and VolMinNet \cite{li2021provably} (denoted as PLM-F, PLM-D, PLM-R, and PLM-V). As both original Forward and T-Revision methods filtered the highest-confidence portions of samples, their methods may not be universally applicable \cite{yao2020dual}. Therefore, to assess PLM's performance in a more general context when combined with various transition matrices, we utilize the highest probability samples in each category as anchors for Forward, T-Revision, PLM-F, and PLM-R. For PLM-R, we also integrated the reweighting strategy outlined in T-Revision \cite{xia2019anchor}, detailed in Appendix A. Table \ref{tab:accuracy-10} illustrates that our approach attains increased accuracy while significantly reducing biases. The results suggest that PLM can serve as a component to enhance the robustness of existing classifier-consistent methods. Furthermore, experiments for transition matrix in Appendix G indicate that PLM can help reduce the matrix estimation error.


\section{Conclusion and limitation}
Estimating the noisy class posterior accurately is critical for noisy label learning. In this paper, we introduces a part-level multi-labeling method aimed at augmenting supervised information, thereby reducing the estimation error of estimating noisy class posterior. By introducing a single-to-multiple transition matrix, we incorporate the part-level supervised information derived from cropped instances into a classifier-consistent framework, effectively mitigating overemphasis. Extensive experiments validate the robust performance of our method, both for estimating the noisy class posterior and for noisy label learning as a component of loss correction. One significant limitation of this study is that cropping original instances with a uniform criterion for labeling may be too simplistic. Future work will involve a detailed exploration of the theories and practices associated with instance representation learning, with the objective of formulating a more appropriate cropping criterion for noisy label learning.

\smallskip\noindent\textbf{Acknowledgments.} This research was partially supported by the National Key Research and Development Project of China No.2021ZD0110700, the Key Research and Development Project in Shaanxi Province No.2023GXLH-024, the National Science Foundation of China under Grant No.62037001 and No.62250009.
{
    \small
    \bibliographystyle{ieeenat_fullname}
    \bibliography{main}
}


\end{document}